# Depression Detection at the Point of Care: Automated Analysis of Linguistic Signals from Routine Primary Care Encounters

Feng Chen[1*], Manas Bedmutha[2*], Janice Sabin[1], Andrea Hartzler[1],
Nadir Weibel[2], Trevor Cohen[1]
* representing equal contributions
[1] University of Washington, Seattle, WA, [2] University of California San Diego, La Jolla, CA

**Abstract**
*Depression is underdiagnosed in primary care, yet timely identification remains critical. Recorded clinical encounters, increasingly common with digital scribing technologies, present an opportunity to detect depression from naturalistic dialogue. We investigated automated depression detection from 1,108 audio-recorded primary care encounters in the Establishing Focus study, with depression defined by PHQ-9 (n=253 depressed, n=855 non-depressed). We compared three supervised approaches, Sentence-BERT + Logistic Regression (LR), LIWC+LR and ModernBERT, against a zero-shot GPT-OSS. GPT-OSS achieved the strongest performance (AUPRC=0.510, AUROC=0.774), with LIWC+LR competitive among supervised models (AUPRC=0.500, AUROC=0.742). Combined dyadic transcripts outperformed single-speaker configurations, with providers linguistically mirroring patients in depression encounters, an additive signal not captured by either speaker alone. Meaningful detection is achievable from the first 128 patient tokens (AUPRC=0.356, AUROC=0.675), supporting in-the-moment clinical decision support. These findings argue for passively collected clinical audio as a low-burden complement to existing screening workflows.*

**Introduction**
Depression is among the most prevalent and burdensome mental health conditions worldwide, affecting an estimated 290 million people globally as of 2019 and projected to become the leading source of disease burden by 2030 [1]. Patients with severe depression have higher suicide risk [2], increased risk of substance abuse, and worse outcomes in chronic disease [3]. However, effective treatments for depression are available, and administering them early can improve disease outcomes [4]. Timely identification of depressive symptoms is therefore critical for effective management of this debilitating condition.

Despite its prevalence, depression remains substantially underdetected in primary care, a setting in which it is frequently encountered [5]. Primary care providers (PCPs) are often the first point of contact for any health concern. The Patient Health Questionnaire (PHQ-9), a widely used screening instrument for depression symptoms, relies on self-reports for longitudinal state over two weeks, both of which are beyond the providers' direct line of visibility. However, providers may not be able to explicitly recognize symptoms of depression in their conversations with patients, with diagnostic sensitivity amongst PCPs estimated at approximately 50% [5], unless the patients express a related complaint. Furthermore, diagnosticians often rely upon disclosure from patients who may be hesitant to disclose information due to social stigma, concerns about loss of autonomy, and lack of trust [6].

To mitigate underdiagnosis, the US Preventive Services Task Force (USPSTF) recommends screening all adults for depression [7]. Most clinics now use check-in software before visits (e.g. VitalSign6[8]) or reminders for health screening [9]. These check-ins require patients to fill the PHQ-2 (an abbreviated two-item depression screening scale) followed by a full PHQ-9 based on the screening scores. These surveys and reminders add to the patient burden and increase survey fatigue[10]. Additionally, in many communities, mental illness is heavily stigmatized therefore patients may be reluctant to disclose depression-related information [11]. As the number of check-ins and checkpoints increases, the perceived quality of care decreases, as administrative burden displaces time and attention from the clinical interaction itself [12]. Therefore methods that can recognize depression-related cues in passively-collected data present a desirable alternative.

Various computational models have shown signal in detecting depression from patient-generated data sources. For example, studies have shown smartphone or passive behavioral data [13], wearables [14] and even social media language patterns to hold indications of depression [15]. While sensor data tracks movement and physiological signals over time, the language data provided by social media provides additional information since the linguistic style of depressed patients has been shown to differ systematically from their counterparts [16]. However, all of these sources operate outside the clinical encounter, meaning the primary care provider may have no visibility into these signals at the point of care, where opportunity to intervene is most immediate.

Recent work closer to clinic visits has explored detecting depressive signals from clinical language, mainly using structured psychiatric interview transcripts to classify depression from patient speech [17]. Research on therapist-patient language in text-based counseling conversations and psychotherapy sessions, has shown that linguistic features of these interactions are associated with clinical outcomes [18,19]. Beyond conversational data, researchers have leveraged EHR clinical notes to identify depression and related mental health risk at the population level, with studies demonstrating that combining unstructured narrative notes with structured EHR data substantially improves predictive accuracy over structured data alone [20,21]. However, these approaches share a common limitation: they operate on data generated within dedicated mental health contexts, structured diagnostic interviews, psychotherapy transcripts, or retrospective clinical documentation, where depression is the explicit focus and linguistic signals are relatively salient. Applying NLP to routine primary care encounters presents a fundamentally different challenge: providers are simultaneously managing multiple competing clinical tasks, depression is rarely the presenting concern, and the relevant linguistic signal must be extracted from unscripted, wide-ranging dialogue rather than symptom-focused interviews [22,23]. To our knowledge, no prior work has applied ASR and NLP to automatically detect depressive signals from naturalistic, unscripted primary care audio recordings, the precise clinical context where real-time identification could most directly inform provider action.

Prior work has emphasized the underdiagnosis of depression in primary care visits and its acute downstream impact on depression management [24,25]. Automatic detection of linguistic markers of depression from a patient's conversation with their provider could potentially improve screening, and support delivery of measurement-based care (MBC) during the visit itself [26]. Prior work on patient-provider communication has shown that conversational style shifts over the course of a visit [22] and that both parties dynamically respond to each other's social and affective signals [23,27]. Motivated by this evidence, we investigate the following research questions:
1) Can primary care conversations provide any signal towards screening depression?
2) Do patient and provider conversation behaviors vary in encounters with patients experiencing depression?
3) How early in a visit do linguistic indications of depression emerge, and when might a provider first be alerted?
4) What are the conversational indicators that distinguish a visit with a potentially depressed patient?

By situating depression detection within the primary care encounter, this work bridges computational mental health modeling and clinical informatics. We argue that conversational signals captured during the visit represent an underexplored yet clinically actionable data source, which may enable timely, workflow-integrated mental health support without increasing burden on patients or providers, both of whom are already navigating competing demands within a time-constrained encounter.

## Methods
### Datasets
This study utilized data from the Establishing Focus (EF) study, a randomized controlled trial conducted between 2002 and 2006 to evaluate a brief intervention for improving physician communication skills at 12 community-based primary care clinics in the Puget Sound region [28,29]. The dataset includes 1,282 real-world, audio-recorded patient-provider encounters; we retained only those with both transcribable audio and a valid PHQ-9 response, yielding a final analytic dataset of N=1,108 visits. The depression labels were derived based on the PHQ-9 total scores with a screening cutoff set at 10, which is common practice [30]. This resulted in 253 patients with positive diagnosis and 855 patients who did not show clinical screening levels of depression.

### Transcript Preparation
We transcribed all (n=1108) the raw audio recordings and separated conversational speakers (diarized) using WhisperX. We also had manual transcripts for a small subset (n=92), which was only used to estimate the quality of transcription. All speakers identified in the transcripts were assigned roles based on three categories: "patient", "doctor", and "others" (typically nurses or research coordinators), using recent BERT-based role classification models developed on manual transcripts from the same dataset [31]. Evaluating over the manual transcripts, we found an average word error rate (WER) of 45.2% and character error rate (CER) of 31.0%; which are likely overestimated due to de-identification placeholders, non-speech annotations in manual transcripts, and poor audio quality. The role detection model achieves 95% accuracy on accurately diarized data.

Across the 1,108 encounters, the mean transcript length was 2,374 tokens (standard deviation [SD]=1,214), with patient-only speech averaging 1,034 tokens (SD =647) and provider-only speech averaging 1,254 tokens (SD =776). For our analysis, "doctor" and "others" were aggregated into a single category (i.e., "provider").

**Model Development and Comparison**

We used three supervised transformer based approaches along with an LLM to detect depression from clinical transcripts. Clinical transcripts are lengthy by nature, as evident in our prepared transcripts as well. Therefore, we chose models that are capable of handling long-form clinical text. Models used in our analysis are described below.

*Sentence-BERT (SBERT) + Logistic Regression (LR):* We generated sentence embeddings using the Sentence-BERT model[32]. Since transcripts frequently exceeded standard BERT token limits, we segmented each transcript into 128-token chunks, and aggregated chunk-level embeddings into a single document representation by mean pooling. These embeddings were fed into a LR classifier (liblinear solver) with balanced class weights.
*Linguistic Inquiry and Word Count (LIWC) + LR:* We extracted psycholinguistic features using LIWC-22 [33] and trained a LR classifier on the resulting feature vectors. Since these descriptors are easily human-interpretable, they also help us characterize depressive states better.
*ModernBERT:* We fine-tuned ModernBERT-base, a long-context encoder-only model optimized for long sequences [34], with a maximum sequence length of 4,096 tokens, batch size of 2, learning rate of $2\times10^{-6}$, trained for 5 epochs with weighted Cross Entropy Loss to penalize minority-class misclassifications.
*GPT (Zero-shot LLM Baseline):* To benchmark against a state-of-the-art generative approach, we evaluated GPT-OSS: 120B [35], a leading open-weight model, in a zero-shot setting. Open-weight deployment allowed secure local hosting of clinical data without transmission to external APIs. The model was prompted to take the role of an experienced psychiatrist and estimate the probability that a patient is at high risk for depression (1.0) or low risk (0.0), outputting only a decimal score. No training examples or labels were provided. This comparison directly addresses whether the linguistic patterns driving supervised model performance are recoverable through general clinical reasoning alone, without task-specific fine-tuning.

All supervised models were evaluated using stratified 5-fold cross-validation across three speaker configurations: full transcripts, patient-only, and provider-only text. To examine the feasibility of early detection, we additionally evaluated SBERT+LR, LIWC+LR and GPT-OSS on truncated transcripts comprising the first 128, 256, and 512 tokens under each speaker configuration.

**Evaluation metrics**

Given the significant class imbalance between depression (n=253) and non-depression (n=855) classes, we report five complementary metrics. AUPRC and AUROC serve as our primary threshold-free metrics: AUPRC captures model performance specifically on the minority (depression) class across all decision thresholds, while AUROC provides a more familiar measure of overall discriminative ability. For threshold-dependent metrics, we selected the decision threshold maximizing F1 score within each cross-validation fold. We then report Balanced Accuracy (BA), Precision (positive predictive value), and Recall (sensitivity) at this threshold. BA ensures equal weight to both classes, while precision and recall characterize the clinical tradeoff between false alarms and missed cases. For the zero-shot GPT baseline, which was not cross-validated, all metrics are reported on the full dataset at the F1-maximizing threshold.

**Linguistic Characterization of Depression**

In addition to understanding the ability to screen for depression, we also aim to understand linguistic markers of depression exhibited in primary-care encounters. We used LIWC features to identify significant differences in language observed in visits for patients with and without depression. LIWC operates by matching words in the transcripts against a validated dictionary of grammatical, psychological, and social categories (e.g., negative emotion, personal pronouns), calculating the percentage of total words that fall into each category. We conducted two-sample t-tests separately for combined, patient-only and provider-only subsets of transcripts across the depression and non-depression groups. We report feature groups with adjusted p<0.05. Additionally, we also use these descriptors to understand which linguistic features are deemed important by the predictive logistic regression model and report the top-10 most indicative features.

## Results
### Overall model performance

Table 1 presents the performance of all four models on full combined dyadic transcripts. GPT-OSS achieved the highest overall performance (AUPRC = 0.510, AUROC = 0.774, BA = 0.704), followed closely by LIWC+LR (AUPRC = 0.500 ± 0.061, AUROC = 0.742 ± 0.035). SBERT+LR achieved an AUPRC of 0.458 (± 0.078) and AUROC of 0.740 (± 0.048), outperforming ModernBERT (AUPRC = 0.394 ± 0.061, AUROC = 0.664 ± 0.049). Notably, LIWC+LR matched SBERT+LR on balanced accuracy (0.693 vs 0.698) while achieving higher AUPRC, and its performance was within 0.01 AUROC points of GPT-OSS despite requiring no neural computation.

**Table 1.** Overall performance of all four models on full combined dyadic transcripts. AUPRC = area under the precision-recall curve; AUROC = area under the receiver operating characteristic curve; BA = balanced accuracy; Precision = positive predictive value (PPV); Recall = sensitivity. Metrics for supervised models are reported as mean ± SD across 5-fold cross-validation. GPT-OSS metrics are reported on the full dataset at the F1-maximizing threshold.

| Model | AUPRC | AUROC | BA | Precision | Recall |
|---|---|---|---|---|---|
| ModernBERT | 0.394 ± 0.061 | 0.664 ± 0.049 | 0.655 ± 0.038 | 0.348 ± 0.058 | **0.756 ± 0.175** |
| SBERT + LR | 0.458 ± 0.078 | 0.740 ± 0.048 | 0.698 ± 0.037 | 0.421 ± 0.092 | 0.727 ± 0.135 |
| LIWC + LR | 0.500 ± 0.061 | 0.742 ± 0.035 | 0.693 ± 0.016 | **0.445 ± 0.100** | 0.684 ± 0.183 |
| GPT-OSS | **0.510** | **0.774** | **0.704** | 0.398 | 0.739 |

### Speaker Configuration Effects

Table 2 presents model performance stratified by speaker configuration. Given ModernBERT's lower performance and high computational cost, speaker-stratified and temporal analyses were conducted on SBERT+LR, LIWC+LR, and GPT-OSS only.

For SBERT+LR and GPT-OSS, patient-only and provider-only configurations performed comparably, both slightly below combined dyadic transcripts performance, suggesting that patient and provider speech carry overlapping but complementary signals. LIWC+LR showed a markedly different pattern: performance dropped substantially when restricted to patient-only (AUPRC = 0.278 ± 0.042) or provider-only transcripts (AUPRC = 0.255 ± 0.039) compared to combined (AUPRC = 0.500 ± 0.061). This gap indicates that LIWC+LR depends on the joint linguistic signal from both speakers.

**Table 2.** Model performance stratified by speaker configuration.

| Model | Speaker Config | AUPRC | AUROC | BA | Precision | Recall |
|---|---|---|---|---|---|---|
| SBERT+LR | Patient-only | 0.425 ± 0.079 | 0.720 ± 0.052 | 0.683 ± 0.031 | 0.376 ± 0.060 | 0.759 ± 0.071 |
| | Provider-only | 0.431 ± 0.057 | 0.711 ± 0.046 | 0.678 ± 0.033 | 0.432 ± 0.036 | 0.589 ± 0.125 |
| | Combined | 0.458 ± 0.078 | 0.740 ± 0.048 | 0.698 ± 0.037 | 0.421 ± 0.092 | 0.727 ± 0.135 |
| LIWC + LR | Patient-only | 0.278 ± 0.042 | 0.557 ± 0.052 | 0.588 ± 0.028 | 0.297 ± 0.055 | 0.753 ± 0.212 |
| | Provider-only | 0.255 ± 0.039 | 0.534 ± 0.070 | 0.563 ± 0.053 | 0.270 ± 0.041 | 0.839 ± 0.150 |
| | Combined | 0.500 ± 0.061 | 0.742 ± 0.035 | 0.693 ± 0.016 | 0.445 ± 0.100 | 0.684 ± 0.183 |
| GPT-OSS | Patient-only | 0.483 | 0.736 | 0.686 | 0.437 | 0.601 |
| | Provider-only | 0.476 | 0.743 | 0.682 | 0.377 | 0.711 |
| | Combined | 0.510 | 0.774 | 0.704 | 0.398 | 0.739 |

### Temporal dynamics of depression signals

Figure 1 presents SBERT+LR, LIWC+LR, and GPT-OSS performance at truncated transcript lengths across speaker configurations. Across SBERT+LR and GPT-OSS, patient-only speech yielded the strongest early signal. At 128 tokens, SBERT+LR patient-only achieved AUPRC = 0.331 (± 0.043) and AUROC = 0.603 (± 0.066), outperforming both provider-only (AUPRC = 0.260 ± 0.022, AUROC = 0.546 ± 0.019) and combined (AUPRC = 0.275 ± 0.038, AUROC = 0.541 ± 0.042) configurations. GPT-OSS showed consistently stronger absolute early performance, with patient-only reaching AUPRC = 0.356 and AUROC = 0.675 at 128 tokens compared to provider-only (AUPRC = 0.280, AUROC = 0.587) and combined (AUPRC = 0.293, AUROC = 0.611).

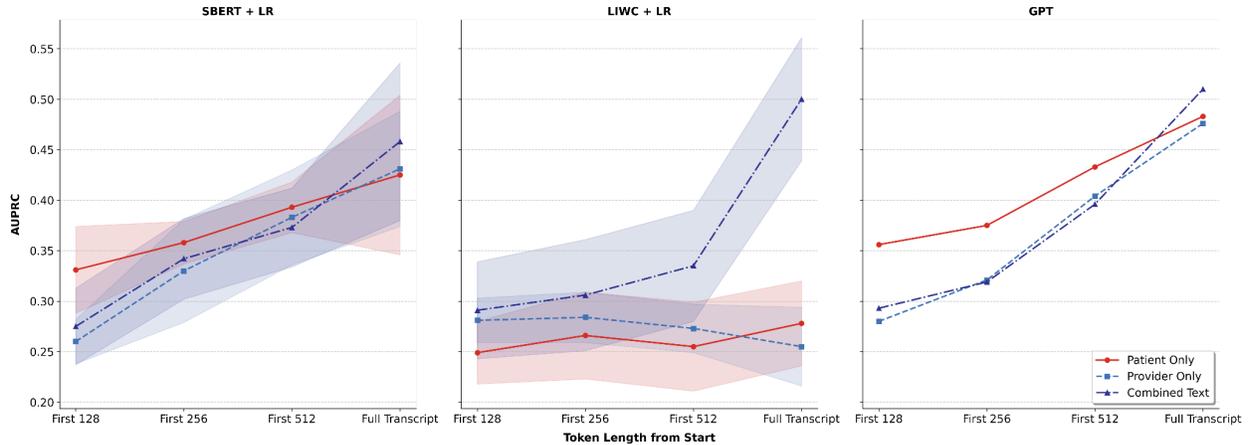

**Figure 1.** AUPRC across token length from start (first 128, 256, and 512 tokens, and full transcripts) for three speaker configurations: patient-only, provider-only, and combined dyadic transcripts. Left panel shows SBERT+LR; middle panel shows LIWC+LR; right panel shows GPT. Shaded bands in the left and middle panels indicate ± 1SD across cross-validation folds.

Provider speech accumulated predictive signal more gradually across both models: SBERT+LR provider-only required approximately 256 tokens (AUPRC = 0.330 ± 0.051) to match the signal detectable in just the first 128 patient tokens, and GPT-OSS showed the same asymmetry (0.321 vs 0.356 at equivalent thresholds). Performance across speaker configurations converges as full transcripts are used for SBERT+LR (AUPRC 0.425–0.458), while GPT-OSS maintains a consistent advantage at every truncation level, suggesting its zero-shot clinical reasoning extracts depressive signal more efficiently regardless of context size.

LIWC+LR showed a markedly different temporal pattern. Unlike SBERT+LR and GPT-OSS, LIWC+LR remained low predictive performance at all truncation levels for both patient-only (AUPRC = 0.249 ± 0.031 at 128 tokens) and provider-only (AUPRC = 0.281 ± 0.022) configurations, with the combined configuration only marginally better (AUPRC = 0.291 ± 0.048). Performance improved only modestly with additional tokens across all configurations, never approaching the full-transcript combined performance (AUPRC = 0.500). This confirms that LIWC+LR's strong full-transcript performance is not simply a function of transcript length, but depends on the richness of the joint dyadic signal.

**LIWC Features analysis**
To characterize linguistic differences associated with depression, we conducted two-sample t-tests across three speaker configurations (combined, patient-only, provider-only) and examined feature weights from the LIWC+LR model trained on combined dyadic transcripts (Figure 2). The combined dyadic transcripts yielded 54 significantly different features, compared to 20 for patient-only and 8 for provider-only (all $p < 0.05$), with the top features for each configuration summarized in Table 4. The significant features provide a linguistic basis for the performance pattern observed in Table 2, where LIWC+LR performance improved markedly with combined transcripts while reduced significantly on single-speaker configurations.

In the combined dyadic transcripts, the strongest group differences were in overall emotional tone (t = 7.53), pronoun categories (t = −6.54), positive tone (t = 5.79), negative emotion (t = −5.74), and sadness-specific language (t = −5.66), with depressed encounters characterized by lower emotional tone, more self-referential pronouns, less positive sentiment, and more negative and sad emotional words. Complementarily, the LIWC+LR model coefficient plot (Figure 2), reflecting predictive weights learned from cross-validation rather than direct group comparisons, largely converges with these findings: emo_sad, mental, home, memory, and substances carried the strongest positive weights toward depression, while tone_pos, family, want, leisure, and allnone predicted non-depression.

In patient-only transcripts, the top five significant features were positive tone, cognition, affect, first-person singular pronouns, and authenticity. Depressed patients used less positive tone and overall affect, and more first-person singular pronouns, cognitive processing language, and authenticity — indicating more spontaneous, less socially filtered speech. In provider-only transcripts, eight features reached significance. Providers in depression encounters used more first-person singular pronouns and total pronouns, more substance-related language, and more present-focused language, while using fewer temporal references and quantitative terms (numbers). Notably, the provider pronoun elevation mirrors the patient pattern observed in both the patient-only and combined analyses.

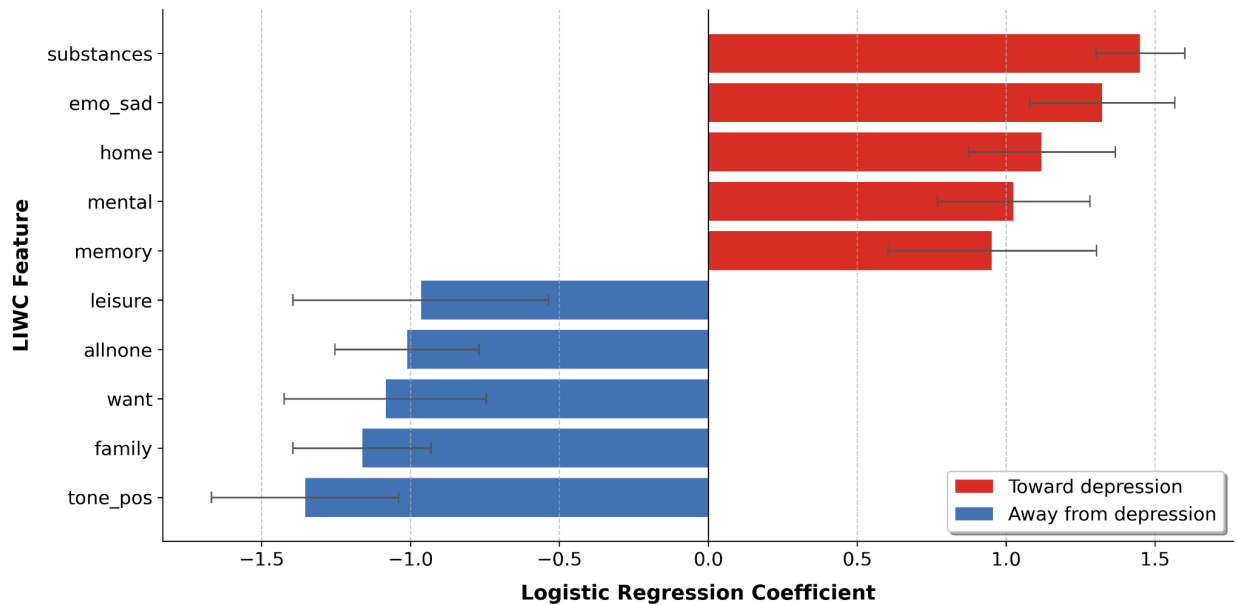

**Figure 2.** Logistic regression coefficients for the top 10 LIWC-22 features from the LIWC+LR model trained on combined dyadic transcripts. Red bars indicate features associated toward depression; blue bars indicate features associated away from depression. Error bars represent ± 1 SD across 5-fold cross-validation folds.

**Table 4.** LIWC-22 features comparisons between depression and non-depression groups by speaker configuration. For each feature, group means are reported for the non-depression (n=855) and depression (n=253) groups, along with the t-statistic from a two-sample t-test. Negative t-statistics indicate higher values in the depression group. All features shown are statistically significant at p < 0.05. Features are ordered by absolute t-statistic.

| Speaker Role | LIWC Feature | Explanation (supporting literatures) | Non-depression group mean | Depression group mean | T-statistic |
|---|---|---|---|---|---|
| **Combined** | Tone | Composite emotional tone (positive minus negative sentiment) | 46.93 | 39.19 | 7.53 |
| | Pronouns categories* | I, it, they, he, she, etc* | 5.59* | 6.27* | -6.54* |
| | Positive Tone | Positive sentiment words | 2.86 | 2.45 | 5.79 |
| | Negative Emotion | Words expressing negative emotions (e,g: sad, angry, nervous) | 0.58 | 0.75 | -5.74 |
| | Sad Emotion | Sadness-specific words (e.g: cry, grief, sorrow) | 0.02 | 0.06 | -5.66 |
| **Patient** | Positive Tone | Positive emotional language [36] | 13.20 | 12.85 | 3.27 |
| | Cognition | Causal/analytical reasoning [36] | 28.67 | 29.30 | -3.19 |
| | Affect | Overall emotional language [37] | 4.48 | 4.12 | 3.16 |
| | I | Self-referential language [16] | 6.53 | 6.92 | -3.04 |
| | Authenticity | Spontaneous, unfiltered expression [38] | 78.58 | 81.72 | -3.01 |
| **Provider** | I | Self-referential language | 5.52 | 5.80 | -2.44 |
| | Time | Temporal references | 5.35 | 5.10 | 2.32 |
| | Substances | Drug/alcohol references | 0.06 | 0.09 | -2.32 |
| | Numbers | Quantitative language | 2.14 | 1.94 | 2.27 |
| | Pronouns | All pronouns combined | 21.66 | 21.97 | -2.12 |

\* The Pronoun categories row consolidates three LIWC features: 1st person singular, general Pronouns and personal pronouns, all of which ranked among the top features in the combined analysis with near-identical t-statistics (−6.54 to −6.36). The mean and t-statistic shown are for 1st Person Singular as the most theoretically specific measure. All three were significant at p < 0.001.

**Discussion**

This study demonstrates the feasibility of detecting depressive symptoms from routine primary care audio recordings using ASR and NLP, and characterizes the linguistic mechanisms underlying that detection across supervised and zero-shot approaches. Among supervised models, SBERT+LR outperformed ModernBERT (AUPRC 0.458 vs. 0.394) despite ModernBERT's long-context design. This suggests that mean pooling over 128-token chunks captures these localized depression symptom disclosures effectively, whereas long-context models may suffer from signal dilution across lengthy, topic-varied clinical dialogue. This aligns with findings by Chen et al. in PTSD detection, where the same architectural pattern holds [39]. Notably, LIWC+LR achieved the highest AUPRC among supervised models (0.500 ± 0.061) and matched SBERT+LR on balanced accuracy, despite relying solely on interpretable psycholinguistic features without any neural representation — a surprising result that underscores the diagnostic salience of established linguistic categories for depression in this setting. GPT-OSS achieved the strongest overall performance (AUPRC = 0.510, AUROC = 0.774), demonstrating that zero-shot clinical reasoning from a large open-weight model can detect depressive signals without task-specific training, and with stronger discriminative ability than all supervised approaches. To contextualize these results clinically, a large meta-analysis of the accuracy of routine clinical diagnosis of depression in primary care found sensitivity of ~50% and PPV of 42% at a reference prevalence of 21.9% [5]. The EF dataset prevalence of 22.8% is consistent with this estimate: given that approximately 50% of depressive episodes go undetected in primary care [40], a diagnosed rate of ~11% implies a true underlying prevalence of approximately 22%, suggesting EF captures a realistic representation of depression burden in routine clinical visits. Our models on combined dyadic transcripts match or exceed the PPV of routine clinical diagnosis (40–45%) while substantially improving sensitivity (68–74%), suggesting that conversation-based screening could complement rather than replace the clinical encounter by flagging cases that unassisted clinical judgment is most likely to miss.

The speaker-stratified analysis reveals that LIWC+LR's strong full-transcript performance depends critically on the combined dyadic signal: it collapsed from AUPRC 0.500 on combined to 0.278 and 0.255 on patient-only and provider-only transcripts respectively, while SBERT+LR and GPT-OSS showed only modest degradation. The LIWC feature analysis explains this: the combined dyadic transcripts yielded 54 significantly different features compared to 20 (patient-only) and 8 (provider-only) meaning that the signals are additive. Notably, providers in depression encounters elevated their use of first-person singular pronouns, mirroring the patient pattern, alongside more present-focused and substance-related language and fewer temporal and quantitative references. This provider linguistic shift, away from structured data-gathering toward a more exploratory register, has to our knowledge not been previously documented using LIWC in clinical encounters, and may reflect an unconscious accommodation to patient status, consistent with evidence that clinicians adapt their function word use, including pronoun patterns, toward patients in clinical encounters [41].

These temporal findings point to the potential for in-the-moment clinical decision support: alerting providers to depressive signals early enough within the visit to steer the encounter toward formal assessment before the clinical window closes. Both SBERT+LR and GPT-OSS yield meaningful signals from the first 128 patient tokens (AUPRC 0.331 and 0.356, AUROC 0.603 and 0.675 respectively). GPT-OSS maintains a consistent advantage at every truncation level, suggesting its zero-shot clinical prediction is more robust regardless of context size. While performance at truncated lengths is lower than with the full transcripts, this tradeoff may be acceptable given the clinical benefit of early alerting, and should be transparently communicated to users of any deployed system. Previous work shows that physicians interrupt patients within the first 11–23 seconds of their opening statements [42,43]. Our results quantify the diagnostic cost of that practice: the opening patient signal is precisely what drives the strongest early detection, and allowing patients an uninterrupted opening preserves it.

The LIWC analysis converges with the temporal findings on a coherent depressive linguistic phenotype. Across the combined dyadic transcripts, the strongest group-level markers were reduced emotional tone, elevated self-referential pronouns, reduced positive sentiment, and increased negative and sadness-specific language, which is consistent with prior depression language research [16,36–38]. Taken together, the convergence of model performance, speaker-stratified analyses, temporal dynamics, and linguistic feature patterns paints a consistent picture: depressive symptoms leave a trackable imprint on clinical dialogue, concentrated in patient speech but amplified when provider accommodation is also captured. These findings argue for treating passively collected clinical audio as an underutilized source of decision support signal, one that could complement existing screening workflows without adding burden to patients or providers navigating an already time-constrained encounter.

**Limitations and Future Work**
This study has several limitations. First, the clinical encounters were recorded between 2002 and 2006, which may limit the generalizability of conversational dynamics to contemporary settings. Second, although PHQ-9 provides a validated measure of depression symptoms, it was administered within the clinical visit itself, meaning ground truth labels may reflect momentary self-disclosure rather than a stable diagnosis. Third, the zero-shot GPT-OSS evaluation used a single prompt and a single open-weight model; we did not systematically explore alternative prompting strategies, prompt formulations, or other LLM architectures. Performance may vary substantially across models and prompt designs. Finally, both ASR transcription errors and residual speaker role misclassification may introduce noise into the analyses. Future work should validate the identified linguistic phenotype prospectively across diverse patient populations and clinical settings. Incorporating acoustic and prosodic features, such as speech rate, pause duration, and vocal affect, could further enrich the depressive signal beyond lexical content. Extending the framework to predict depression severity rather than binary classification, and ultimately integrating the proposed lightweight pipeline into a prospective clinical decision support evaluation, represent important next steps toward real-world deployment.

**Conclusion**
This study demonstrates that depressive symptoms leave a detectable linguistic imprint in routine primary care encounters, one that emerges early in patient speech, is amplified by provider accommodation, and is recoverable through automated speech recognition and natural language processing. GPT-OSS achieved the strongest overall performance (AUPRC = 0.510), demonstrating that a large open-weight model can detect depressive signals without task-specific fine-tuning. Among supervised models, LIWC+LR matched GPT-OSS on AUPRC (0.500 ± 0.061) and outperformed long-context ModernBERT (0.394), suggesting that interpretable psycholinguistic features carry diagnostic signal comparable to neural representations in this setting. The first 128 patient tokens alone yield an AUPRC of 0.356 from GPT-OSS, and combined dyadic transcripts outperform either speaker alone, with providers linguistically mirroring patients in depression encounters. The combined configuration yielded 54 significantly differentiated LIWC features compared to 20 and 8 for patient-only and provider-only respectively, with providers mirroring patients by elevating first-person singular pronouns and substance-related language in depression encounters. Lexical analyses converge on a coherent depressive linguistic phenotype: elevated self-referential pronouns, increased sadness and mental health language, and reduced positive emotional tone. These findings argue that passively collected clinical audio represents a low-burden data source for in-the-moment decision support, with the potential to complement existing screening workflows without adding burden to patients or providers.

**Acknowledgements**
We would like to thank Dr Lynne S. Robins and the team for developing the Establishing Focus dataset. And we would also like to thank Andrew Zolensky for developing and sharing the speaker identification model used to assign clinical roles to diarized speaker labels in this study.